\documentclass{mosi}
\usepackage[utf8]{inputenc}
\usepackage{array,makecell,threeparttable,adjustbox}
\usepackage{algorithm,algorithmic}
\usepackage{xurl}
\setcitestyle{authoryear,round,citesep={;},aysep={,},yysep={;}}
\hypersetup{pdftitle={MARCO: Multi-Round Agentic Reinforcement for Conditional Molecular Optimization},pdfauthor={Shicheng Fang, Yuxin Wang, Zhuo Yang, Xiaohu Xu, Jiahao Lu, Chuanyuan Tan, Tong Zhu, Yining Zheng, Xipeng Qiu}}
\newcommand{\up}{$\uparrow$}
\newcommand{\down}{$\downarrow$}
\newcommand{\bnum}[1]{\textbf{#1}}
\newcommand{\marco}{MARCO}
\newcommand{\repo}{RePO}
\newcommand{\grpo}{GRPO}
\newcommand{\sftinit}{\textsuperscript{*}}
\newcommand{\tabrepo}{RePO}
\newcommand{\tabgrpo}{GRPO}
\newcommand{\tabmarco}{MARCO}
\newcommand{\tabgrpostar}{GRPO\sftinit{}}
\newcommand{\tabmarcostar}{MARCO\sftinit{}}
\newcommand{\qwenThreeB}{Qwen2.5-3B-Instruct}
\newcommand{\qwenFourB}{Qwen3-4B-Instruct-2507}
\newcommand{\qwenSevenB}{Qwen2.5-7B-Instruct}
\newcommand{\mistralSevenB}{Mistral-7B-Instruct-v0.3}
\newcommand{\gellmoPSixMistral}{\ensuremath{\mathrm{GeLLM}^{3}\mathrm{O}\text{-}\mathrm{P}(6)_{\mathrm{Mistral}}}}
\newcommand{\qwenThreeBHead}{\makecell{\textbf{Qwen2.5-3B}\\\textbf{Instruct}}}
\newcommand{\qwenFourBHead}{\makecell{\textbf{Qwen3-4B}\\\textbf{Instruct-2507}}}
\newcommand{\qwenSevenBHead}{\makecell{\textbf{Qwen2.5-7B}\\\textbf{Instruct}}}

\newtcolorbox{supplementtext}[1]{
    enhanced,
    colframe=black!55,
    colback=black!2,
    colbacktitle=black!5,
    coltitle=black,
    fonttitle=\bfseries,
    title={#1},
    boxrule=0.55pt,
    arc=1.5pt,
    left=6pt,
    right=6pt,
    top=5pt,
    bottom=5pt,
    before skip=4pt,
    after skip=4pt
}

\title{\marco: Multi-Round Agentic Reinforcement\\for Conditional Molecular Optimization}
\author{Shicheng Fang$^{1,2}$, Yuxin Wang$^{1,2,\dagger}$, Zhuo Yang$^{2,4}$, Xiaohu Xu$^{2,5}$, Jiahao Lu$^{1,2}$, Chuanyuan~Tan$^{3}$, Tong Zhu$^{2,5}$, Yining Zheng$^{1,2}$, Xipeng Qiu$^{1,2,\dagger}$}
\affil[1]{Fudan University}
\affil[2]{Shanghai Innovation Institute}
\affil[3]{Soochow University}
\affil[4]{Southeast University}
\affil[5]{East China Normal University}
\authormark{$^{\dagger}$Corresponding author}

\abstract{Molecular optimization is inherently iterative: a candidate is proposed, evaluated against several objectives, and revised while preserving a relationship to the source molecule. 
Most instruction-following models instead emit one edited molecule, forcing validity, property improvement, and similarity control into a single response. 
We introduce \marco, an evaluator-grounded reinforcement-learning framework that trains molecular editors on bounded proposal--feedback--revision trajectories. 
MARCO aggregates shaped turn rewards into an undiscounted trajectory return for group-relative policy optimization. 
We evaluate two consequences of this training: Same-1 tests the trained policy under a one-response budget, while Same-5 tests whether the same policy can use verifier feedback when up to five responses are available. 
Across the three-objective MuMOInstruct benchmark, three Qwen backbones, and seen/unseen instruction splits, SFT-initialized MARCO obtains the highest product of property success rate and similarity in every reported primary setting. Same-5 further improves the observed score under the tested budget, while four-objective and public-checkpoint experiments test transfer across constraint sets and initialization regimes.}
\checkdata[Code]{\url{https://github.com/euReKa025/MARCO-release}}
\begin{document}
\raggedbottom
\maketitle

\section{Introduction}

Conditional molecular optimization is a closed-loop task: the candidate is proposed, evaluated against multiple objectives, and revised according to the remaining deficits. Existing instruction-following pipelines usually train and evaluate a one-shot mapping, so one response must satisfy validity, property improvement, and source-structure similarity without a structured opportunity for correction.

Recent generative models, instruction-tuning datasets, and LLM-based systems improve molecule generation from natural-language objectives or reference examples \cite{tang2024surveygenerativeainovo,li2024tomg,doi:10.1021/acs.jcim.5c02234,dey-etal-2025-gellm3o,ye2025drugassist,li2026reference}. The central mismatch remains: the task is iterative, but the dominant policy interface is one-shot. We study whether exposing a policy to verifier-guided revision during training improves its first response, and whether the same policy can continue using feedback when additional inference turns are affordable.

We propose \marco, short for Multi-Round Agentic Reinforcement for Conditional Molecular Optimization. Here, \emph{agentic} denotes only a bounded policy--verifier interaction: the model observes structured feedback and revises its next molecular action. The verifier checks validity, directional property changes, and source-molecule fingerprint similarity at each turn. MARCO starts from a per-subtask SFT checkpoint, aggregates shaped turn rewards into a trajectory return, and applies a group-relative policy update. Same-1 therefore tests the first response after trajectory training; Same-5 tests the same policy with verifier feedback during inference. MARCO makes three contributions:
\begin{itemize}
    \item It casts instruction-based molecular optimization as a bounded feedback-conditioned decision process and trains a molecular editor from verifier-grounded trajectory returns that combine property progress, structural similarity, validity, and revision quality.
    \item It evaluates the same trained policy at two operating points: Same-1 tests transfer to a first response under a single-response budget, while Same-5 tests revision when additional verifier-guided responses are available.
    \item It provides ablation and transfer evidence across initialization, training horizon, objective count, and public-checkpoint adaptation, clarifying where the complete MARCO recipe is effective.
\end{itemize}

\section{Related Work}

\paragraph{Classical and neural molecular optimization.}
Molecular optimization has been studied through graph-based generation, fragment editing, reinforcement learning, and goal-directed search \cite{jin2019junctiontreevariationalautoencoder,olivecrona2017molecularnovodesigndeep,you2018gcpn,chen2021deep,fu2021mimosa,gao2022sampleefficiency}. Some methods modify molecular graphs or fragments directly, while sequence-to-sequence approaches such as Chemformer and PMol cast optimization as translation over SMILES strings learned from molecule pairs \cite{irwin2022chemformer,wu2024leveraging}. Evolutionary approaches, including recent LLM-assisted search over chemical space, provide another path for proposing improved candidates \cite{wang2025efficient}. These methods establish the property--similarity tradeoff, but do not provide a common language-model training interface for verifier-guided revision.

\paragraph{Training-free LLM methods.}
One line of LLM-based chemistry work uses prompting, retrieval, tools, or dialogue without updating the underlying policy. Speak-to-Structure evaluates open-domain natural-language-driven molecule generation \cite{li2024tomg}. MuMOInstruct formulates source-conditioned instruction-following multi-property optimization \cite{dey-etal-2025-gellm3o}. Tool- and dialogue-based systems such as ChemCrow, autonomous chemistry agents, ChatDrug, and Re3DF further show how retrieval, domain feedback, and iterative interaction support chemistry-facing workflows at inference time \cite{m2024augmenting,boiko2023autonomous,liu2024conversational,le-etal-2025-agentdrug}. These systems expose interaction at inference time, but do not use the resulting verifier trajectories to update the molecular editor.

\paragraph{Training-based LLM methods.}
Another line updates the model through instruction tuning or reinforcement learning. DrugAssist instruction-tunes LLMs for molecular optimization, while Mol-Instructions, LlaSMol, and GeLLM$^3$O develop chemistry or biomolecular instruction resources \cite{ye2025drugassist,fang2024molinstructions,yu2024llasmol,dey-etal-2025-gellm3o}. PPO and GRPO-style objectives provide widely used foundations for language-model post-training with verifiable rewards \cite{schulman2017proximal,shao2024deepseekmath,guo2025deepseek}. MolAct uses a two-stage editing-to-optimization curriculum with multi-turn, tool-augmented RL on ChemCoTBench \cite{yang2025molact}. RePO applies reference-guided single-turn policy optimization to molecular editing \cite{li2026reference}. Multi-turn RL has also been studied for language-model agents operating in interactive environments \cite{zhou2024archer,wang2025ragen}. MARCO uses verifier-grounded trajectories for source-conditioned multi-property optimization and evaluates the same trained policy under one- and five-response budgets.

\section{Problem Formulation}

Let \(x_0 \in \mathcal{X}\) be a source molecule represented as a SMILES string \cite{weininger1988smiles}. An instruction specifies an active property set \(\mathcal{P}\) and a direction vector \(\mathbf{d}=\{d_p\}_{p\in\mathcal{P}}\), where \(d_p\in\{+1,-1\}\) indicates whether property \(p\) should increase or decrease. Each property has a scoring function \(f_p:\mathcal{X}\rightarrow\mathbb{R}\). For a candidate molecule \(x\), the directional improvement for property \(p\) is
\begin{equation}
\Delta_p(x;x_0) = d_p \cdot \bigl(f_p(x) - f_p(x_0)\bigr).
\end{equation}
A candidate succeeds on the requested properties when every active directional improvement is positive:
\begin{equation}
\mathrm{Succ}(x;x_0,\mathcal{P},\mathbf{d}) =
\mathbf{1}\left[\forall p \in \mathcal{P},\; \Delta_p(x;x_0)>0\right].
\end{equation}
Auxiliary properties may be logged for analysis, but they are not part of the active success or reward target unless included in \(\mathcal{P}\).

The molecule must also preserve a useful structural relationship to the source. We use Tanimoto similarity over Morgan fingerprints \cite{rogers2010extended}, a common similarity measure in cheminformatics \cite{bajusz2015tanimoto,lopez2024molecular}:
\begin{equation}
s(x,x_0)=
\frac{|\mathrm{FP}(x)\cap \mathrm{FP}(x_0)|}
{|\mathrm{FP}(x)\cup \mathrm{FP}(x_0)|}.
\end{equation}
Validity is treated separately from property success. A reported candidate must parse as a valid single connected molecule. We denote this event by \(V(x)=1\). Similarity acceptance uses lower and upper fingerprint-similarity bounds:
\begin{equation}
A_{\mathrm{sim}}(x;x_0)=
\mathbf{1}\left[\delta_{\mathrm{low}} \le s(x,x_0) < \delta_{\mathrm{high}}\right].
\end{equation}
The lower bound limits fingerprint deviation from the source, and the upper bound prevents near-identity candidates from receiving full editing credit.

In single-turn optimization, a policy directly produces one candidate \(\hat{x}\) for the instruction. In \marco{} training, the same instance is lifted to a bounded \(T\)-turn decision process. The policy can observe feedback from previous proposals before choosing the next candidate, but evaluation still reports a selected final candidate under the specified interaction budget.

\section{Method}

\begin{figure}[!t]
    \centering
    \includegraphics[width=\textwidth]{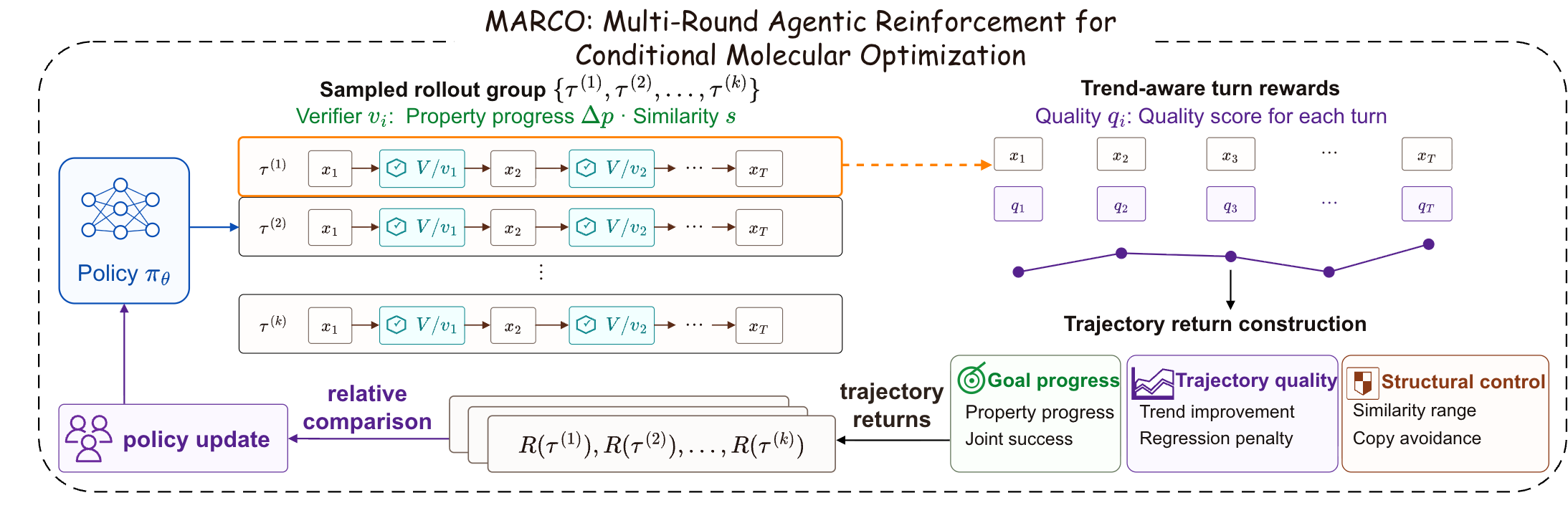}
    \caption{Overview of \marco{} trajectory learning. The policy samples bounded trajectories in which candidate molecules alternate with verifier feedback on validity, directional property progress, and source similarity. After rollout, the shaped turn rewards are aggregated into one trajectory return; group-relative comparison of these returns updates the policy.}
    \label{fig:marco_overview}
\end{figure}

\marco{} turns conditional molecular optimization into bounded feedback-conditioned trajectory learning. Figure~\ref{fig:marco_overview} summarizes the RL stage. For each instruction, the policy samples a group of trajectories that alternate between candidate molecules and verifier feedback on validity, directional property progress, and source similarity. The realized shaped turn rewards are aggregated into one trajectory return. Returns are compared within the rollout group to update the policy. In the main recipe, this RL stage starts from a per-subtask SFT policy that provides initial molecular-editing behavior.

Intuitively, a proposal should move the requested properties in the correct directions and remain within a useful similarity range rather than drift or copy the source. Later attempts also receive credit when they improve after verifier feedback. Invalid proposals receive a fixed penalty, but the trajectory continues so later valid actions can still be evaluated.

\subsection{Environment Feedback}

For an instruction \(u=(x_0,\mathcal{P},\mathbf{d})\), the initial history is \(h_1=u\). At turn \(t\), the policy samples a molecular response
\begin{equation}
o_t \sim \pi_{\theta}(\cdot\mid h_t),
\qquad
x_t = \psi(o_t)\in\mathcal{X}\cup\{\bot\},
\end{equation}
where \(\psi\) extracts the proposed molecule and returns the sentinel \(\bot\) when the response does not yield a valid, evaluable candidate. The environment feedback is
\begin{equation}
z_t=
\begin{cases}
\bigl(1,\{\Delta_p(x_t;x_0)\}_{p\in\mathcal{P}},
s_t,A_{\mathrm{sim},t},m_t\bigr), & x_t\in\mathcal{X},\\
\bigl(0,m_t^{\mathrm{inv}}\bigr), & x_t=\bot,
\end{cases}
\end{equation}
where \(s_t=s(x_t,x_0)\), \(m_t\) is valid-candidate feedback, and \(m_t^{\mathrm{inv}}\) reports the invalid action. Property deltas, similarity, and valid-candidate quality are not evaluated for \(\bot\). The next history is
\begin{equation}
h_{t+1}=h_t \oplus (o_t,z_t).
\end{equation}
The realized trajectory is \(\tau=(h_1,o_1,z_1,\ldots,o_{T_\tau},z_{T_\tau})\), truncated early when a proposal satisfies both property success and similarity acceptance.

Invalid format, invalid molecule parse, disconnected molecules, and unevaluable candidates are treated as failed model actions. An invalid action receives invalid feedback and the trajectory continues.

\subsection{Trajectory Return Construction}

For a valid candidate, property quality is a clipped directional-progress score:
\begin{equation}
Q_{\mathrm{prop}}(x_t)=
\frac{1}{|\mathcal{P}|}
\sum_{p\in\mathcal{P}}
\operatorname{clip}\!\left(\Delta_p(x_t;x_0),0,c_p\right),
\end{equation}
where \(c_p\) is a property-specific clipping scale. This term measures only directional property progress; structural preservation is handled by the separate similarity-quality term below.

Similarity quality is range-aware:
\begin{equation}
Q_{\mathrm{sim}}(s_t)=
\begin{cases}
-\alpha_{\mathrm{low}}\dfrac{\delta_{\mathrm{low}}-s_t}{\delta_{\mathrm{low}}}, & s_t < \delta_{\mathrm{low}},\\[6pt]
\dfrac{s_t-\delta_{\mathrm{low}}}{\delta_{\mathrm{high}}-\delta_{\mathrm{low}}}, & \delta_{\mathrm{low}} \le s_t < \delta_{\mathrm{high}},\\[6pt]
1-\alpha_{\mathrm{copy}}\dfrac{s_t-\delta_{\mathrm{high}}}{1-\delta_{\mathrm{high}}}, & s_t \ge \delta_{\mathrm{high}}.
\end{cases}
\end{equation}
This reward shape favors the accepted fingerprint-similarity range while penalizing low-similarity edits and near-copy outputs.

The per-turn state quality is
\begin{equation}
\begin{aligned}
q_t &=
w_{\mathrm{prop}} Q_{\mathrm{prop}}(x_t)
+
w_{\mathrm{sim}} Q_{\mathrm{sim}}(s_t)
\\
&\quad+
b_{\mathrm{succ}}\,
\mathrm{Succ}(x_t;x_0,\mathcal{P},\mathbf{d})\\
&\quad\cdot
A_{\mathrm{sim}}(x_t;x_0),
\end{aligned}
\end{equation}
with a gated bonus when both property and similarity conditions hold. The trend contribution is
\begin{equation}
\phi_t=
\begin{cases}
0, & t=1,\\[3pt]
\begin{aligned}
&\lambda_{\mathrm{imp}}
\left[q_t-\max_{j<t}q_j-\eta_{\mathrm{imp}}\right]_+\\[-1pt]
&\quad-\lambda_{\mathrm{reg}}
\left[q_{t-1}-q_t-\eta_{\mathrm{reg}}\right]_+
\end{aligned} & t\ge 2.
\end{cases}
\end{equation}
The trend terms are zero at the first turn. The shaped turn reward is
\begin{equation}
r_t=
\begin{cases}
r_{\mathrm{inv}}, & x_t=\bot,\\[3pt]
q_t+\phi_t-\lambda_{\mathrm{len}}\mathbf{1}[\ell_t>\ell_{\max}], & x_t\in\mathcal{X},
\end{cases}
\end{equation}
where \(r_{\mathrm{inv}}<0\) is the fixed invalid-action reward, \([a]_+=\max(a,0)\), \(\ell_t\) is the generated length, and \(\eta_{\mathrm{imp}}\) and \(\eta_{\mathrm{reg}}\) are trend margins. For an invalid model action, \marco{} assigns the fixed invalid-action penalty and carries forward the previous state-quality value for trajectory bookkeeping; the invalid turn contributes no improvement or regression term. Subsequent valid turns compare their quality against this carried-forward history. Let \(r_t\) denote the shaped reward at turn \(t\). We use the undiscounted trajectory return
\begin{equation}
R(\tau)=\sum_{t=1}^{T_{\tau}} r_t.
\end{equation}
Unexecuted turns lie outside the realized trajectory and do not appear in the sum.

\subsection{Policy Optimization}

For each instruction \(u\), \marco{} samples a group of \(K\) trajectories \(\{\tau_i\}_{i=1}^K\). Each trajectory receives a scalar return \(R_i=R(\tau_i)\). Following group-relative policy optimization \cite{shao2024deepseekmath}, the advantage is
\begin{equation}
A_i = \frac{R_i - \mu(R_{1:K})}{\sigma(R_{1:K}) + \epsilon}.
\end{equation}
Let \(\mathcal{M}_i\) be the set of model-generated tokens in trajectory \(\tau_i\), excluding environment feedback tokens. The clipped PPO-style objective \cite{schulman2017proximal} is
\begin{equation}
\begin{aligned}
\mathcal{J}_{\mathrm{MARCO}}(\theta)
&=
\mathbb{E}_{u,\{\tau_i\}_{i=1}^K}
\Bigl[
\frac{1}{K}\sum_{i=1}^{K}\bar{L}_i
\\
&\qquad -
\beta\,\mathbb{D}_{\mathrm{KL}}(\pi_{\theta}\Vert\pi_{\mathrm{ref}})
\Bigr],\\
\bar{L}_i
&=
\frac{1}{|\mathcal{M}_i|}
\sum_{k\in\mathcal{M}_i}
\min\Bigl(
\rho_{i,k}A_i,
\\
&\qquad
\operatorname{clip}(\rho_{i,k},1-\varepsilon,1+\varepsilon)A_i
\Bigr).
\end{aligned}
\end{equation}
where \(\rho_{i,k}\) is the policy ratio for token \(k\). Masking the environment feedback text is important: the update is applied to the model's molecular-editing actions, not to the verifier messages that the environment inserts between turns.

\section{Experiments}

\subsection{Experimental Setup}
\paragraph{Benchmark}
We evaluate on MuMOInstruct, introduced by GeLLM$^3$O for instruction-following multi-property molecule optimization \cite{dey-etal-2025-gellm3o}. The primary benchmark uses three-objective BDP, BDQ, and BPQ tasks across \qwenThreeB, \qwenSevenB, and \qwenFourB{} backbones \cite{qwen2025qwen25technicalreport,yang2025qwen3}. Following MuMOInstruct, seen instructions use phrasings and property names observed during instruction tuning, whereas unseen instructions use the held-out paraphrase and alternative property names. We also report HMPQ and BDPQ as four-objective extensions, with Same-1 rows in the main text and Same-5 details in Supplementary Material, Sec. D.4.
Following MuMOInstruct, we use its released benchmark property evaluators. BBBP, DRD2, HIA, and mutagenicity are learned scoring-oracle outputs in the goal-directed benchmark sense \cite{brown2019guacamol}, whereas QED \cite{Bickerton_2012} and penalized logP are deterministic molecular scores. Table~\ref{tab:tasks} states every requested optimization direction. For experiments newly run for this paper, each reported value is the arithmetic mean over three independent runs. Prior-paper data retain the published values.

\begin{table}[!htbp]
    \centering
    \begin{minipage}{.52\textwidth}\centering
    \small
    \begin{tabular*}{\linewidth}{@{\extracolsep{\fill}}ll@{}}
        \toprule
        \textbf{Task} & \textbf{Requested directions} \\
        \midrule
        BDP & BBBP\up{} + DRD2\up{} + pLogP\up{} \\
        BDQ & BBBP\up{} + DRD2\up{} + QED\up{} \\
        BPQ & BBBP\up{} + pLogP\up{} + QED\up{} \\
        HMPQ & HIA\up{} + mutagenicity\down{} + pLogP\up{} + QED\up{} \\
        BDPQ & BBBP\up{} + DRD2\up{} + pLogP\up{} + QED\up{} \\
        \bottomrule
    \end{tabular*}
    \caption{Molecular optimization tasks and requested directions. BDP, BDQ, and BPQ form the primary three-objective benchmark; HMPQ and BDPQ are four-objective extensions. pLogP denotes penalized logP.}
    \label{tab:tasks}
    \end{minipage}
\end{table}

\paragraph{Baselines and training.}

We compare the base instruction model, per-subtask SFT, \grpo{}, \repo{}, and \marco{}. In prose, we refer to SFT-initialized and base-initialized configurations explicitly. In compact tables, a superscript \sftinit{} marks initialization from the corresponding per-subtask SFT checkpoint. 
Following the original RePO recipe, we initialize RePO from the base instruction model and retain its answer-level reference guidance during policy optimization \cite{li2026reference}. The primary RePO--MARCO\textsuperscript{*} comparison therefore evaluates their complete training recipes.

The \grpo{} baseline follows the RePO-compatible one-response reward: candidates outside the lower similarity gate receive zero, while accepted candidates combine source similarity with the fraction of improved active properties \cite{li2026reference}. The full expression is given in Supplementary Material, Sec. B.1.

We use two evaluation budgets. Same-1 denotes single-response same-harness evaluation and is the primary protocol. Same-5 is an auxiliary interaction-budget extension that permits up to five responses; within this evaluation, all compared methods receive the same verifier-feedback fields, accumulated interaction history, parsing rules, and candidate-selection procedure. Same-5 is reported across \qwenThreeB, \qwenSevenB, and \qwenFourB. Supplementary Material documents the four-objective runs and the \gellmoPSixMistral{} public-checkpoint adaptation experiment based on \mistralSevenB{}.
All reported evaluations cover the complete seen or unseen benchmark split without evaluation-time subsampling.

\newcommand{\primaryresulttable}{%
\begin{table}[!t]
    \centering
    \small
    \setlength{\tabcolsep}{0.75pt}
    \renewcommand{\arraystretch}{1.02}
    \begin{threeparttable}
    \begin{adjustbox}{max width=\linewidth}
    \begin{tabular}{@{}llcccccccccccccccccc@{}}
        \toprule
        \textbf{Metric} & \textbf{Method} & \multicolumn{6}{c}{\qwenThreeBHead} & \multicolumn{6}{c}{\qwenSevenBHead} & \multicolumn{6}{c}{\qwenFourBHead} \\
        \cmidrule(lr){3-8} \cmidrule(lr){9-14} \cmidrule(lr){15-20}
        & & \multicolumn{3}{c}{\textbf{Seen}} & \multicolumn{3}{c}{\textbf{Unseen}} & \multicolumn{3}{c}{\textbf{Seen}} & \multicolumn{3}{c}{\textbf{Unseen}} & \multicolumn{3}{c}{\textbf{Seen}} & \multicolumn{3}{c}{\textbf{Unseen}} \\
        \cmidrule(lr){3-5} \cmidrule(lr){6-8} \cmidrule(lr){9-11} \cmidrule(lr){12-14} \cmidrule(lr){15-17} \cmidrule(lr){18-20}
        & & BDP & BDQ & BPQ & BDP & BDQ & BPQ & BDP & BDQ & BPQ & BDP & BDQ & BPQ & BDP & BDQ & BPQ & BDP & BDQ & BPQ \\
        \midrule
        \multirow{5}{*}{SR} & Base & 0.052 & 0.034 & 0.052 & 0.052 & 0.042 & 0.050 & 0.098 & 0.082 & 0.134 & 0.074 & 0.088 & 0.254 & 0.130 & 0.120 & 0.154 & 0.118 & 0.162 & 0.172 \\
        & SFT & \underline{0.398} & \bnum{0.319} & \bnum{0.471} & \underline{0.310} & \bnum{0.342} & \underline{0.419} & \underline{0.368} & \underline{0.290} & \underline{0.522} & \bnum{0.408} & \underline{0.334} & \underline{0.546} & \underline{0.450} & \bnum{0.284} & \underline{0.532} & \underline{0.436} & \underline{0.274} & \underline{0.558} \\
        & \tabgrpo & 0.156 & 0.082 & 0.212 & 0.148 & 0.078 & 0.186 & 0.068 & 0.074 & 0.158 & 0.070 & 0.124 & 0.180 & 0.000 & 0.000 & 0.290 & 0.000 & 0.000 & 0.292 \\
        & \tabrepo & 0.206 & 0.160 & 0.274 & 0.198 & 0.170 & 0.242 & 0.276 & 0.170 & 0.258 & 0.302 & 0.250 & 0.292 & 0.158 & 0.248 & 0.300 & 0.272 & 0.210 & 0.356 \\
        & \tabmarcostar & \bnum{0.418} & \underline{0.288} & \underline{0.440} & \bnum{0.378} & \underline{0.274} & \bnum{0.512} & \bnum{0.370} & \bnum{0.382} & \bnum{0.588} & \underline{0.402} & \bnum{0.372} & \bnum{0.580} & \bnum{0.498} & \underline{0.282} & \bnum{0.570} & \bnum{0.466} & \bnum{0.284} & \bnum{0.594} \\
        \cmidrule(lr){1-20}
        \multirow{5}{*}{Sim} & Base & 0.149 & 0.117 & 0.194 & 0.143 & 0.104 & 0.130 & \underline{0.630} & \underline{0.686} & \underline{0.660} & \underline{0.591} & 0.631 & \underline{0.535} & \underline{0.671} & 0.617 & \underline{0.641} & \underline{0.649} & \underline{0.690} & \underline{0.698} \\
        & SFT & 0.254 & 0.279 & 0.244 & 0.261 & 0.257 & 0.248 & 0.481 & 0.670 & 0.482 & 0.465 & \underline{0.639} & 0.446 & 0.559 & 0.644 & 0.525 & 0.559 & 0.656 & 0.500 \\
        & \tabgrpo & \bnum{0.759} & \underline{0.479} & \bnum{0.567} & \bnum{0.727} & \underline{0.457} & \underline{0.573} & \bnum{0.850} & \bnum{0.818} & \bnum{0.798} & \bnum{0.822} & \bnum{0.777} & \bnum{0.759} & \bnum{0.998} & \bnum{1.000} & \bnum{0.873} & \bnum{0.998} & \bnum{1.000} & \bnum{0.870} \\
        & \tabrepo & \underline{0.569} & 0.365 & 0.509 & \underline{0.572} & 0.322 & \bnum{0.596} & 0.435 & 0.352 & 0.448 & 0.344 & 0.301 & 0.403 & 0.656 & \underline{0.675} & 0.621 & 0.608 & 0.680 & 0.599 \\
        & \tabmarcostar & 0.485 & \bnum{0.576} & \underline{0.527} & 0.464 & \bnum{0.570} & 0.523 & 0.529 & 0.573 & 0.527 & 0.501 & 0.591 & 0.533 & 0.541 & 0.652 & 0.579 & 0.542 & 0.656 & 0.531 \\
        \cmidrule(lr){1-20}
        \multirow{5}{*}{SR$\times$Sim} & Base & 0.008 & 0.004 & 0.010 & 0.007 & 0.004 & 0.007 & 0.062 & 0.056 & 0.088 & 0.044 & 0.055 & 0.136 & 0.087 & 0.074 & 0.099 & 0.077 & 0.112 & 0.120 \\
        & SFT & 0.101 & \underline{0.089} & 0.115 & 0.081 & \underline{0.088} & 0.104 & \underline{0.177} & \underline{0.194} & \underline{0.252} & \underline{0.190} & \underline{0.213} & \underline{0.244} & \underline{0.252} & \underline{0.183} & \underline{0.280} & \underline{0.244} & \underline{0.180} & \underline{0.279} \\
        & \tabgrpo & \underline{0.118} & 0.039 & 0.120 & 0.108 & 0.036 & 0.107 & 0.058 & 0.061 & 0.126 & 0.058 & 0.096 & 0.137 & 0.000 & 0.000 & 0.253 & 0.000 & 0.000 & 0.254 \\
        & \tabrepo & 0.117 & 0.058 & \underline{0.139} & \underline{0.113} & 0.055 & \underline{0.144} & 0.120 & 0.060 & 0.116 & 0.104 & 0.075 & 0.118 & 0.104 & 0.167 & 0.186 & 0.165 & 0.143 & 0.213 \\
        & \tabmarcostar & \bnum{0.203} & \bnum{0.166} & \bnum{0.232} & \bnum{0.175} & \bnum{0.156} & \bnum{0.268} & \bnum{0.196} & \bnum{0.219} & \bnum{0.310} & \bnum{0.201} & \bnum{0.220} & \bnum{0.309} & \bnum{0.270} & \bnum{0.184} & \bnum{0.330} & \bnum{0.253} & \bnum{0.186} & \bnum{0.315} \\
        \bottomrule
    \end{tabular}\end{adjustbox}
    \begin{tablenotes}[flushleft]
        \small
        \item[$*$] Initialization from the corresponding per-subtask SFT checkpoint. Best and second-best values within each metric/backbone/split/objective column are bolded and underlined, respectively.
    \end{tablenotes}
    \caption{Primary Same-1 results under single-response same-harness evaluation. Backbones are grouped horizontally; columns report seen/unseen instruction splits and three objectives.}
    \label{tab:same1_main}
    \end{threeparttable}
\end{table}
}

\paragraph{Evaluation protocol and metrics.}
\label{sec:metrics}

To preserve comparability with prior work, the primary Same-1 results use the established MuMOInstruct benchmark and RePO-compatible reporting convention: the official seen/unseen instruction splits with unchanged membership, released property evaluators, property-only success, and the SR$\times$Sim aggregate \cite{dey-etal-2025-gellm3o,li2026reference}.

For an evaluation set with \(N\) instructions, let \(\hat{x}_i\) be the selected candidate under the evaluation budget. In Same-1, \(\hat{x}_i\) is the single generated candidate. In Same-5, \(\hat{x}_i\) is the first property-success candidate if one appears before the horizon and the horizon candidate otherwise. Define
\begin{equation}
\begin{aligned}
v_i &= V(\hat{x}_i),\\
y_i &= v_i\,\mathrm{Succ}(\hat{x}_i;x_{0,i},\mathcal{P}_i,\mathbf{d}_i),\\
s_i &=
\begin{cases}
s(\hat{x}_i,x_{0,i}), & v_i=1,\\
0, & v_i=0.
\end{cases}
\end{aligned}
\end{equation}
Success rate is property-only, following the RePO-compatible benchmark definition \cite{li2026reference}:
\begin{equation}
\mathrm{SR}=\frac{1}{N}\sum_{i=1}^{N} y_i .
\end{equation}
Similarity is the mean selected-candidate Tanimoto similarity over valid selected molecules:
\begin{equation}
\mathrm{Sim}=
\frac{\sum_{i=1}^{N} v_i s_i}
{\max\left(1,\sum_{i=1}^{N}v_i\right)}.
\end{equation}
The benchmark summary multiplies these two aggregate statistics:
\begin{equation}
\mathrm{SR}\times\mathrm{Sim}=\mathrm{SR}\cdot\mathrm{Sim}.
\end{equation}

\primaryresulttable

\subsection{Main Single-Turn Results}

Table~\ref{tab:same1_main} presents the primary Same-1 comparison under a single-response budget. Across all three backbones, both instruction splits, and all three objectives, MARCO achieves the highest SR$\times$Sim in every reported cell. Because Same-1 does not permit inference-time revision, these results evaluate the learned first response without additional verifier calls. The consistent gains therefore show that verifier-guided trajectory training can improve the initial molecular-editing action.

The component metrics show that this advantage is not explained by similarity alone. A policy may obtain high Sim by making negligible changes while failing to improve the requested properties. For example, on Qwen3-4B-Instruct-2507 BDQ, GRPO reaches Sim${}=1.000$ and SR${}=0$ on both instruction splits. MARCO instead retains nonzero property success while preserving substantial source similarity, producing the highest SR$\times$Sim in the same setting. This result reflects a better balance between property optimization and structural preservation rather than an isolated increase in either component.

Because SR and Sim are marginal statistics computed over different candidate sets, we additionally audit success-weighted similarity (SWS) at the candidate level. Let $c_i$ indicate exact molecular identity with the source. SWS retains similarity only for a valid, non-copy candidate that improves every requested property direction:
\begin{equation}
\mathrm{SWS}=\frac{1}{N}\sum_{i=1}^{N}v_i y_i(1-c_i)s_i.
\end{equation}
Using aligned local Same-1 outputs for all methods, MARCO obtains the highest SWS in every audited primary task--split cell for all three backbones (Table~\ref{tab:sws_same1_all}). This candidate-level audit is complementary to the benchmark SR$\times$Sim metric and further checks that the gains are not explained by exact-copy behavior.

\begin{table}[!t]
    \centering
    \small
    \setlength{\tabcolsep}{0.75pt}
    \renewcommand{\arraystretch}{1.02}
    \begin{adjustbox}{max width=\linewidth}
    \begin{tabular}{@{}llcccccccccccccccccc@{}}
        \toprule
        \textbf{Metric} & \textbf{Method} & \multicolumn{6}{c}{\qwenThreeBHead} & \multicolumn{6}{c}{\qwenSevenBHead} & \multicolumn{6}{c}{\qwenFourBHead} \\
        \cmidrule(lr){3-8} \cmidrule(lr){9-14} \cmidrule(lr){15-20}
        & & \multicolumn{3}{c}{\textbf{Seen}} & \multicolumn{3}{c}{\textbf{Unseen}} & \multicolumn{3}{c}{\textbf{Seen}} & \multicolumn{3}{c}{\textbf{Unseen}} & \multicolumn{3}{c}{\textbf{Seen}} & \multicolumn{3}{c}{\textbf{Unseen}} \\
        \cmidrule(lr){3-5} \cmidrule(lr){6-8} \cmidrule(lr){9-11} \cmidrule(lr){12-14} \cmidrule(lr){15-17} \cmidrule(lr){18-20}
        & & BDP & BDQ & BPQ & BDP & BDQ & BPQ & BDP & BDQ & BPQ & BDP & BDQ & BPQ & BDP & BDQ & BPQ & BDP & BDQ & BPQ \\
        \midrule
        \multirow{5}{*}{SWS} & Base & 0.040 & 0.027 & 0.055 & 0.023 & 0.027 & 0.048 & 0.063 & 0.048 & 0.083 & 0.043 & 0.054 & 0.129 & 0.081 & 0.068 & 0.092 & 0.069 & 0.093 & 0.102 \\
        & SFT & 0.162 & 0.149 & 0.205 & 0.147 & 0.133 & 0.241 & 0.170 & 0.186 & 0.247 & 0.178 & 0.200 & 0.246 & 0.239 & 0.169 & 0.281 & 0.231 & 0.181 & 0.268 \\
        & \tabgrpo & 0.056 & 0.068 & 0.119 & 0.042 & 0.065 & 0.119 & 0.049 & 0.046 & 0.098 & 0.048 & 0.087 & 0.106 & 0.000 & 0.000 & 0.223 & 0.000 & 0.000 & 0.217 \\
        & \tabrepo & 0.082 & 0.050 & 0.084 & 0.053 & 0.060 & 0.072 & 0.150 & 0.085 & 0.156 & 0.172 & 0.113 & 0.178 & 0.098 & 0.148 & 0.168 & 0.158 & 0.115 & 0.192 \\
        & \tabmarco & \bnum{0.164} & \bnum{0.154} & \bnum{0.208} & \bnum{0.148} & \bnum{0.137} & \bnum{0.260} & \bnum{0.187} & \bnum{0.201} & \bnum{0.298} & \bnum{0.196} & \bnum{0.203} & \bnum{0.305} & \bnum{0.265} & \bnum{0.171} & \bnum{0.326} & \bnum{0.245} & \bnum{0.184} & \bnum{0.311} \\
        \bottomrule
    \end{tabular}\end{adjustbox}
    \begin{tablenotes}[flushleft]
        \small
        \item[] SWS uses aligned local Same-1 outputs and a fixed complete-split denominator. Bold marks the highest numerical value in each task--split column. The audit is separate from prior-paper aggregate rows retained in the primary comparison.
    \end{tablenotes}
    \caption{Candidate-level success-weighted similarity (SWS) under Same-1. Invalid candidates, property failures, and exact source copies contribute zero.}
    \label{tab:sws_same1_all}
\end{table}

\subsection{Trajectory-Return Audit}

We audited the realized trajectory returns on the validation rollouts used for model selection. Across the audited runs, trajectories that achieved joint property and similarity success generally received higher returns than unsuccessful trajectories, while trajectories that reached the rollout horizon without success had substantially lower mean and median returns. For example, on Qwen3-4B-Instruct-2507 BPQ, early-success trajectories achieved a mean return of 1.905, whereas unsuccessful trajectories that reached the four-turn horizon averaged $-2.988$. This separation provides an empirical sanity check that the shaped trajectory return assigns higher credit to the intended optimization behavior.

\subsection{Ablations}

\paragraph{Initialization.}

The initialization ablation separates the contribution of the supervised editing prior from that of subsequent trajectory training on \qwenThreeB{}. SFT-initialized MARCO achieves the highest SR$\times$Sim in all six objective/split columns, while base-initialized MARCO improves over the corresponding base model in every cell. These comparisons indicate that trajectory training provides measurable gains even without supervised initialization, although its effect is substantially stronger when the policy already possesses task-specific molecular-editing behavior. GRPO also changes under SFT initialization, confirming that the starting policy influences reinforcement-learning outcomes. Nevertheless, the matched rows consistently show that the strongest first-response policy results from combining the supervised molecular-editing prior with MARCO's verifier-guided trajectory training.

\begin{table}[!t]
    \centering
    \small
    \setlength{\tabcolsep}{0.75pt}
    \renewcommand{\arraystretch}{1.02}
    \begin{threeparttable}
    \begin{adjustbox}{max width=\linewidth}
    \begin{tabular}{@{}llcccccccccccccccccc@{}}
        \toprule
        \textbf{Metric} & \textbf{Method} & \multicolumn{6}{c}{\qwenThreeBHead} & \multicolumn{6}{c}{\qwenSevenBHead} & \multicolumn{6}{c}{\qwenFourBHead} \\
        \cmidrule(lr){3-8} \cmidrule(lr){9-14} \cmidrule(lr){15-20}
        & & \multicolumn{3}{c}{\textbf{Seen}} & \multicolumn{3}{c}{\textbf{Unseen}} & \multicolumn{3}{c}{\textbf{Seen}} & \multicolumn{3}{c}{\textbf{Unseen}} & \multicolumn{3}{c}{\textbf{Seen}} & \multicolumn{3}{c}{\textbf{Unseen}} \\
        \cmidrule(lr){3-5} \cmidrule(lr){6-8} \cmidrule(lr){9-11} \cmidrule(lr){12-14} \cmidrule(lr){15-17} \cmidrule(lr){18-20}
        & & BDP & BDQ & BPQ & BDP & BDQ & BPQ & BDP & BDQ & BPQ & BDP & BDQ & BPQ & BDP & BDQ & BPQ & BDP & BDQ & BPQ \\
        \midrule
        \multirow{5}{*}{SR$\times$Sim} & Base & 0.096 & 0.076 & 0.141 & 0.061 & 0.096 & 0.132 & 0.128 & 0.136 & 0.191 & 0.113 & 0.130 & 0.215 & 0.108 & 0.094 & 0.139 & 0.114 & 0.142 & 0.149 \\
        & SFT & \underline{0.226} & 0.191 & \underline{0.289} & \underline{0.198} & \underline{0.183} & \underline{0.307} & 0.185 & 0.225 & 0.282 & 0.198 & 0.238 & \underline{0.263} & 0.259 & 0.193 & 0.280 & \underline{0.256} & 0.184 & 0.281 \\
        & \tabgrpo & 0.091 & 0.066 & 0.159 & 0.095 & 0.071 & 0.155 & 0.105 & 0.079 & 0.220 & 0.121 & 0.099 & 0.213 & 0.002 & 0.000 & \underline{0.296} & 0.000 & 0.000 & \underline{0.283} \\
        & \tabrepo & 0.156 & \underline{0.198} & 0.234 & 0.099 & 0.169 & 0.159 & \underline{0.230} & \underline{0.255} & \underline{0.290} & \underline{0.229} & \underline{0.241} & 0.249 & \underline{0.275} & \underline{0.198} & 0.226 & 0.243 & \underline{0.195} & 0.241 \\
        & \tabmarcostar & \bnum{0.232} & \bnum{0.258} & \bnum{0.358} & \bnum{0.211} & \bnum{0.244} & \bnum{0.369} & \bnum{0.241} & \bnum{0.311} & \bnum{0.375} & \bnum{0.236} & \bnum{0.297} & \bnum{0.367} & \bnum{0.289} & \bnum{0.200} & \bnum{0.336} & \bnum{0.265} & \bnum{0.207} & \bnum{0.318} \\
        \bottomrule
    \end{tabular}\end{adjustbox}
    \begin{tablenotes}[flushleft]
        \small
        \item[$*$] Initialization from the corresponding per-subtask SFT checkpoint. Complete SR and Sim are in Supplementary Material, Sec. D.1; bold and underline mark the best and second-best values.
    \end{tablenotes}
    \caption{Same-5 SR$\times$Sim across backbones, instruction splits, and objectives.}
    \label{tab:same5_srxsim}
    \end{threeparttable}
\end{table}

\paragraph{Interaction budget and training horizon.}

Same-5 tests whether the trained policy can use feedback when more than one response is available. Every method receives the same five-response budget and matched verifier/selection protocol. Table~\ref{tab:same5_srxsim} reports the compact comparison, and Figure~\ref{fig:mt5_gain} visualizes the within-configuration gains over Same-1.

The gain bars are computed for each fixed backbone, objective, and instruction split, and are positive throughout. A matched training-horizon control provides more specific evidence: the five-turn-trained policy improves over the one-turn-trained control in all six task--split cells under both evaluation budgets. Averaged over the six cells, it yields a relative SR$\times$Sim increase of 5.2\% under Same-1 and 21.7\% under Same-5.

\subsection{Qualitative Case Study}

An illustrative five-turn trajectory shows successive proposals that miss different requested property directions before the final candidate jointly
improves BBBP, penalized logP, and QED while retaining a source similarity of 0.627. This example illustrates feedback-conditioned revision at the trajectory level and is not intended to replace the aggregate evaluation. Supplementary Material, Sec. F.1, provides the source molecule, candidate SMILES, RDKit depictions, verifier-feedback context, and raw predictor deltas.

\begin{figure}[!t]
    \centering
    \includegraphics[width=0.86\textwidth]{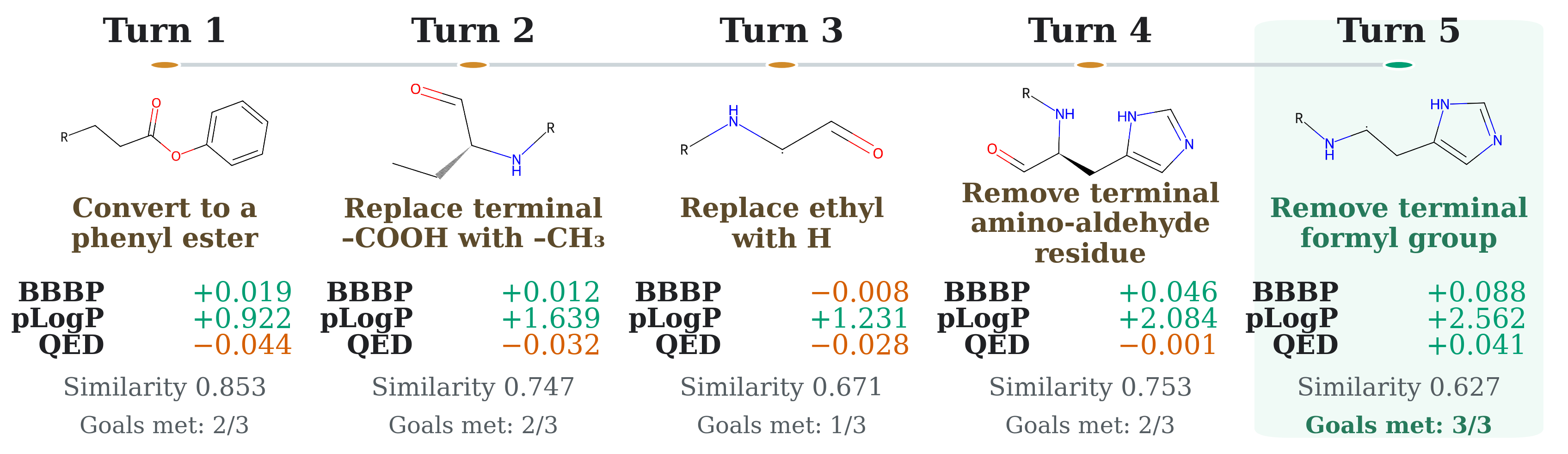}
    \caption{Five-turn BPQ optimization trajectory. Each panel shows the edited local fragment, with $R$ denoting the omitted structure; the Turn-1 label describes the change from the source and later labels describe changes from the preceding candidate. The targets are BBBP$\uparrow$, pLogP$\uparrow$, and QED$\uparrow$, and values are signed changes from the source.}
    \label{fig:case_study_main}
\end{figure}

\begin{table}[!t]
    \centering
    \begin{minipage}{\textwidth}\centering
    \small
    \setlength{\tabcolsep}{4.0pt}
    \renewcommand{\arraystretch}{1.03}
    \begin{tabular*}{\linewidth}{@{\extracolsep{\fill}}llcccccc@{}}
        \toprule
        \textbf{Train horizon} & \textbf{Eval. budget} & \textbf{BDP Seen} & \textbf{BDP Unseen} & \textbf{BDQ Seen} & \textbf{BDQ Unseen} & \textbf{BPQ Seen} & \textbf{BPQ Unseen} \\
        \midrule
        1 turn & Same-1 & 0.184 & 0.165 & 0.158 & 0.142 & 0.225 & 0.267 \\
        5 turns & Same-1 & 0.203 & 0.175 & 0.166 & 0.156 & 0.232 & 0.268 \\
        1 turn & Same-5 & 0.222 & 0.200 & 0.180 & 0.170 & 0.290 & 0.312 \\
        5 turns & Same-5 & 0.232 & 0.211 & 0.258 & 0.244 & 0.358 & 0.369 \\
        \bottomrule
    \end{tabular*}
    \caption{Training-horizon control on Qwen2.5-3B-Instruct MARCO-from-SFT (SR$\times$Sim).}
    \label{tab:training_horizon_control_main}
    \end{minipage}
\end{table}

\subsection{Four-Objective and Transfer Extensions}

HMPQ combines increasing HIA, penalized logP, and QED with decreasing mutagenicity; BDPQ requests four increases. On \qwenSevenB{}, MARCO has the highest Same-1 SR$\times$Sim for both tasks and splits. Its margins over SFT are larger on HMPQ than BDPQ. Same-5 rows and component metrics are in Supplementary Material, Sec. D.4. These extensions increase the number of simultaneous property constraints, so they test whether the recipe transfers beyond the primary three-objective setting. Applying MARCO feedback RL to the released GeLLM$^3$O checkpoint built on Mistral-7B also improves SR$\times$Sim in all six task--split cells under the direct-SMILES five-turn protocol; complete SR, Sim, and SR$\times$Sim results are reported in Table~\ref{tab:p6_full}.

\begin{table}[!htbp]
    \centering
    \small
    \setlength{\tabcolsep}{2.4pt}
    \renewcommand{\arraystretch}{1.05}
    \begin{threeparttable}
    \begin{tabular*}{\linewidth}{@{\extracolsep{\fill}}llcccccc@{}}
        \toprule
        \textbf{Task} & \textbf{Split} & \multicolumn{3}{c}{\textbf{Released checkpoint}} & \multicolumn{3}{c}{\textbf{After feedback RL}} \\
        \cmidrule(lr){3-5} \cmidrule(lr){6-8}
        & & \textbf{SR} & \textbf{Sim} & \textbf{SR$\times$Sim} & \textbf{SR} & \textbf{Sim} & \textbf{SR$\times$Sim} \\
        \midrule
        \multirow{2}{*}{BDP} & Seen & \underline{0.554} & \underline{0.598} & \underline{0.332} & \bnum{0.584} & \bnum{0.614} & \bnum{0.358} \\
        & Unseen & \underline{0.518} & \underline{0.616} & \underline{0.319} & \bnum{0.536} & \bnum{0.630} & \bnum{0.338} \\
        \midrule
        \multirow{2}{*}{BDQ} & Seen & \underline{0.556} & \bnum{0.613} & \underline{0.341} & \bnum{0.578} & \underline{0.605} & \bnum{0.350} \\
        & Unseen & \underline{0.462} & \bnum{0.650} & \underline{0.300} & \bnum{0.500} & \underline{0.637} & \bnum{0.319} \\
        \midrule
        \multirow{2}{*}{BPQ} & Seen & \underline{0.700} & \underline{0.581} & \underline{0.407} & \bnum{0.732} & \bnum{0.594} & \bnum{0.435} \\
        & Unseen & \underline{0.666} & \underline{0.604} & \underline{0.402} & \bnum{0.690} & \bnum{0.614} & \bnum{0.424} \\
        \bottomrule
    \end{tabular*}
    \begin{tablenotes}[flushleft]
        \small
        \item[] Bold and underlined entries mark the better and second value within each metric/task/split pair.
    \end{tablenotes}
    \caption{Public-checkpoint adaptation of \gellmoPSixMistral{} under the direct-SMILES five-turn protocol.}
    \label{tab:p6_full}
    \end{threeparttable}
\end{table}

\begin{figure}[!htbp]
\centering
\begin{minipage}[t]{0.46\textwidth}\vspace{0pt}\centering
\small
    \setlength{\tabcolsep}{2.5pt}
    \renewcommand{\arraystretch}{1.05}
    \begin{threeparttable}
    \begin{tabular*}{\linewidth}{@{\extracolsep{\fill}}lcccccc@{}}
        \toprule
        \textbf{Method} & \multicolumn{3}{c}{\textbf{Seen instruction}} & \multicolumn{3}{c}{\textbf{Unseen instruction}} \\
        \cmidrule(lr){2-4} \cmidrule(lr){5-7}
        & \textbf{BDP} & \textbf{BDQ} & \textbf{BPQ} & \textbf{BDP} & \textbf{BDQ} & \textbf{BPQ} \\
        \midrule
        Base & 0.008 & 0.004 & 0.010 & 0.007 & 0.004 & 0.007 \\
        SFT & 0.101 & 0.089 & 0.115 & 0.081 & 0.088 & 0.104 \\
        \tabgrpo & 0.118 & 0.039 & 0.120 & 0.108 & 0.036 & 0.107 \\
        \tabgrpostar & 0.082 & \underline{0.100} & \underline{0.154} & 0.066 & \underline{0.108} & \underline{0.176} \\
        \tabrepo & 0.117 & 0.058 & 0.139 & \underline{0.113} & 0.055 & 0.144 \\
        \tabmarco & \underline{0.143} & 0.065 & 0.091 & 0.106 & 0.067 & 0.117 \\
        \tabmarcostar & \bnum{0.203} & \bnum{0.166} & \bnum{0.232} & \bnum{0.175} & \bnum{0.156} & \bnum{0.268} \\
        \bottomrule
    \end{tabular*}
    \begin{tablenotes}[flushleft]
        \small
        \item[$*$] Initialization from the corresponding per-subtask SFT checkpoint.
        \item[] Best and second-best values in each column are bolded and underlined, respectively.
    \end{tablenotes}
    \captionof{table}{Initialization ablation on \qwenThreeB{} Same-1. Values are SR$\times$Sim.}
    \label{tab:init_ablation_compact}
    \end{threeparttable}
    
\vspace{12pt}

    \small
    \setlength{\tabcolsep}{2.0pt}
    \renewcommand{\arraystretch}{1.05}
    \begin{threeparttable}
    \begin{tabular*}{\linewidth}{@{\extracolsep{\fill}}lcccc@{}}
        \toprule
        \textbf{Method} & \multicolumn{2}{c}{\textbf{HMPQ}} & \multicolumn{2}{c}{\textbf{BDPQ}} \\
        \cmidrule(lr){2-3} \cmidrule(lr){4-5}
        & \textbf{Seen} & \textbf{Unseen} & \textbf{Seen} & \textbf{Unseen} \\
        \midrule
        Base & 0.026 & 0.037 & 0.016 & 0.027 \\
        SFT & \underline{0.222} & \underline{0.250} & \underline{0.127} & \underline{0.117} \\
        \tabgrpo & 0.028 & 0.014 & 0.054 & 0.031 \\
        \tabrepo & 0.097 & 0.114 & 0.058 & 0.043 \\
        \tabmarcostar & \bnum{0.257} & \bnum{0.285} & \bnum{0.134} & \bnum{0.130} \\
        \bottomrule
    \end{tabular*}
    \begin{tablenotes}[flushleft]
        \small
        \item[$*$] Initialization from the corresponding per-subtask SFT checkpoint.
        \item[] Best and second-best values in each column are bolded and underlined, respectively.
    \end{tablenotes}
    \captionof{table}{Four-objective Same-1 SR$\times$Sim on \qwenSevenB{}; Same-5 results are in Supplementary Material, Sec. D.4.}
    \label{tab:four_obj_main}
    \end{threeparttable}
    \end{minipage}
\hfill
\begin{minipage}[t]{0.49\textwidth}\vspace{0pt}\centering
    \includegraphics[width=0.96\linewidth]{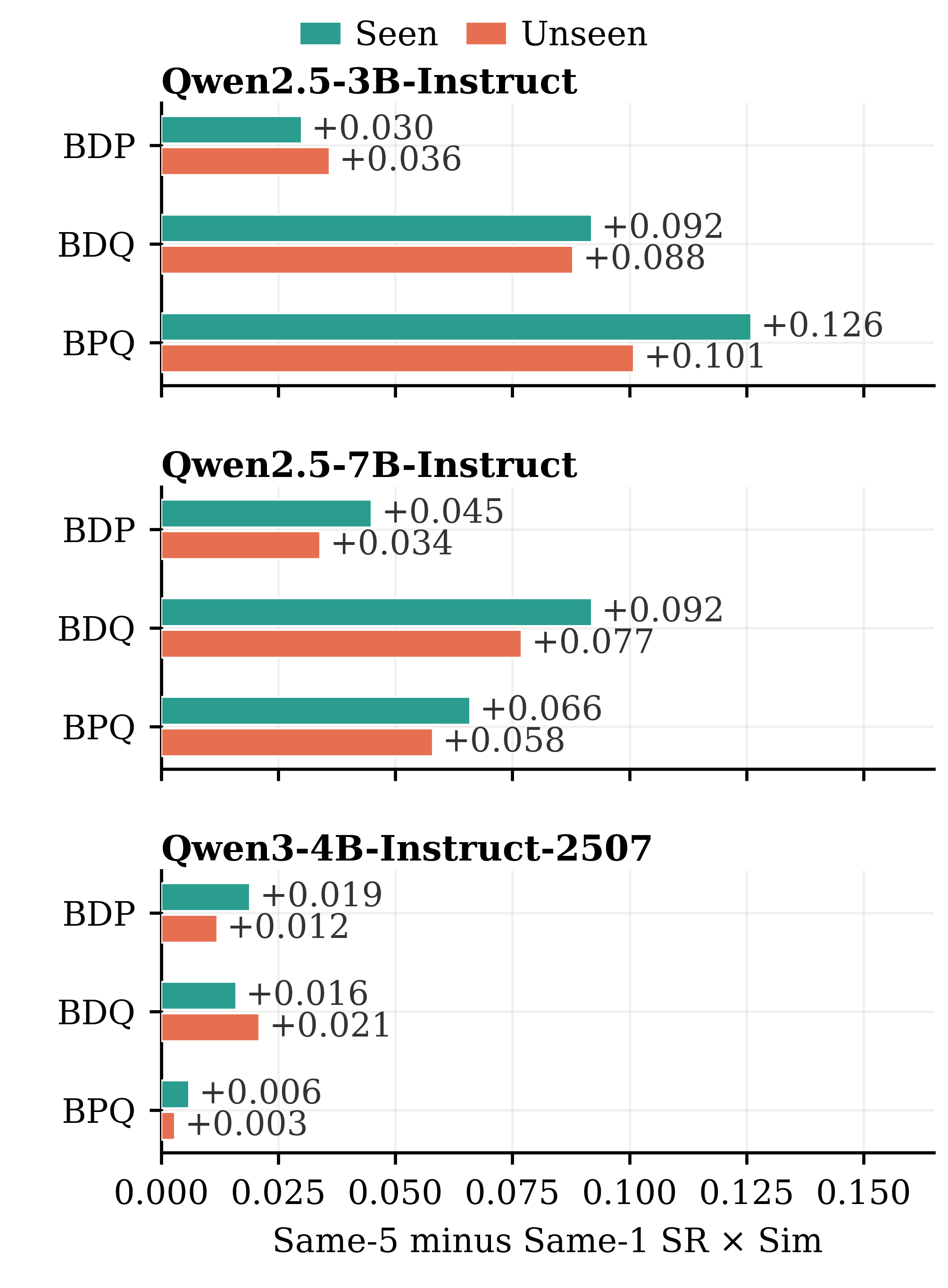}
    \caption{Same-5 minus Same-1 SR$\times$Sim for \marco{}$^*$ by backbone, split, and objective.}
    \label{fig:mt5_gain}
    \end{minipage}
\end{figure}

\FloatBarrier
\section{Conclusion}

\marco{} addresses the mismatch between the iterative nature of molecular optimization and the one-shot training paradigm used by most instruction-following molecular editors. Learning from bounded proposal--feedback--revision trajectories improves the policy's first molecular-editing action under Same-1 while preserving its ability to exploit verifier feedback under Same-5. The training-horizon control suggests that the Same-5 gains are associated with multi-turn training exposure, not only with a larger evaluation budget. Four-objective experiments and public-checkpoint adaptation extend the evaluation to denser constraints and a distinct initialization regime. Together, these results show that trajectory training turns verifier feedback into stronger initial edits while retaining feedback-driven revision. MARCO therefore unifies one-response and bounded interactive molecular optimization in one policy.

\section*{Acknowledgments}
This work is supported by SCION (Scientific Collaborative Innovation with Agentic Organizational Nexus) from Shanghai Innovation Institute.

\clearpage
\bibliographystyle{aaai2027}
\bibliography{main}
\clearpage
\appendix
\appendixpage
% Supplementary content, adapted only for the OpenMOSS single-column layout.

\noindent This supplementary document provides the experimental details needed to interpret the main-paper comparisons. It reports complete ablation metrics, the common evaluation and candidate-selection protocol, policy and verifier-feedback templates, and an illustrative five-turn optimization trajectory.

\section{Experimental Protocol}

\subsection{Tasks and model roles.}
MuMOInstruct BDP, BDQ, and BPQ constitute the three-objective benchmark and are evaluated on seen and unseen instruction splits \cite{dey-etal-2025-gellm3o}. The requested directions are BBBP$\uparrow$/DRD2$\uparrow$/pLogP$\uparrow$ for BDP, BBBP$\uparrow$/DRD2$\uparrow$/QED$\uparrow$ for BDQ, and BBBP$\uparrow$/pLogP$\uparrow$/QED$\uparrow$ for BPQ. Here pLogP denotes penalized logP. The backbone study covers \qwenThreeB{}, \qwenSevenB{}, and \qwenFourB{} under the same task construction. The HMPQ and BDPQ four-objective extensions use \qwenSevenB{}: HMPQ requests HIA$\uparrow$/mutagenicity$\downarrow$/pLogP$\uparrow$/QED$\uparrow$, and BDPQ requests BBBP$\uparrow$/DRD2$\uparrow$/pLogP$\uparrow$/QED$\uparrow$. Initialization ablations use \qwenThreeB{} to study the per-subtask SFT starting point. The public-checkpoint adaptation experiment starts from \gellmoPSixMistral{}, based on \mistralSevenB{} \cite{dey-etal-2025-gellm3o,jiang2023mistral}.

\subsection{Instruction splits.}
We use the official MuMOInstruct seen and unseen instruction splits without changing their membership. Each task provides five diverse instruction phrasings for instruction tuning and one held-out phrasing for unseen evaluation. Unseen instructions also use alternative property names that are absent from instruction tuning.

\subsection{Property evaluators and molecular processing.}
Following MuMOInstruct, we use its released property evaluators \cite{dey-etal-2025-gellm3o}. BBBP, DRD2, HIA, and mutagenicity are predictor scores, whereas QED \cite{Bickerton_2012} and penalized logP are deterministic molecular scores. We compute Tanimoto similarity using 2048-bit radius-2 Morgan bit fingerprints \cite{rogers2010extended} generated by RDKit without chirality encoding \cite{landrum_rdkit}. Molecules are parsed and sanitized with RDKit and must contain a single connected component. We apply no additional tautomer, charge, or salt standardization before fingerprint computation.

\subsection{Comparable evaluation.}
The same-harness protocol uses matched instruction splits, predictor semantics, parsing rules, and candidate selection. Same-1 permits one response and measures feedback-trained behavior under a single-response budget. Same-5 permits up to five responses, selecting the first valid property-success candidate or, if none succeeds, the horizon candidate. Under Same-5, Base, SFT, GRPO, RePO, and MARCO receive the same revision-context template, verifier-feedback fields, and accumulated interaction history at each turn. We report property success rate (SR), source-molecule Tanimoto similarity over valid selected candidates (Sim), and SR$\times$Sim.

The \qwenThreeB{} Same-1 primary-table Base, SFT, RePO and GRPO aggregate values follow RePO \cite{li2026reference}. The \qwenThreeB{} initialization ablation reported below was newly trained and evaluated for this paper. Newly covered backbones, task combinations, and five-turn controls use standalone evaluation under the same protocol.

\section{Training and Evaluation Settings}

Table~\ref{tab:training_evaluation_settings} summarizes the generation budgets, optimization settings, and standalone evaluation conditions used in the reported experiments.

\subsection{Baseline reward definition}

The \grpo{} baseline follows the RePO-compatible one-response reward. Given a proposed molecule $x$, candidates outside the lower similarity gate $G_{\mathrm{sim}}(x;x_0)=\mathbf{1}[s(x,x_0)\geq\delta_{\mathrm{low}}]$ receive zero. Accepted candidates combine source similarity with the fraction of improved active properties:
\begin{equation}
\begin{aligned}
r_{\mathrm{1shot}}(x)
&=\begin{cases}
r_{\mathrm{edit}}(x), & G_{\mathrm{sim}}(x;x_0)=1,\\
0, & G_{\mathrm{sim}}(x;x_0)=0,
\end{cases}\\
r_{\mathrm{edit}}(x)
&=\dfrac{1}{2}s(x,x_0)+\dfrac{1}{2|\mathcal{P}|}
\sum_{p\in\mathcal{P}}\mathbf{1}\left[\Delta_p(x;x_0)>0\right].
\end{aligned}
\end{equation}

\begin{table}[!t]
    \centering
    \small
    \setlength{\tabcolsep}{5.0pt}
    \renewcommand{\arraystretch}{1.03}
    \begin{tabular*}{\linewidth}{@{\extracolsep{\fill}}llll@{}}
        \toprule
        \multicolumn{2}{c}{\textbf{Evaluation and SFT}} & \multicolumn{2}{c}{\textbf{Feedback RL and reward}} \\
        \cmidrule(lr){1-2}\cmidrule(lr){3-4}
        Maximum new tokens & 1024 & Initialization & Per-subtask SFT policy \\
        Inference batch size & 8 & Rollout horizon & 5 turns \\
        Decoding & Deterministic & Training batch / group size $K$ & 32 / 8 \\
        SFT sequence length & 4096 & Response length & 1024 \\
        SFT epochs & 2 & Actor learning rate & $1\times10^{-6}$ \\
        SFT learning rate & $5\times10^{-6}$ & Temperature / top-$p$ & 1.0 / 0.8 \\
        Effective batch, 3B / 7B & 32 / 16 & KL coefficient & 0.05 \\
        Micro-batch, 3B / 7B & 2 / 1 per GPU & Screening interval & Every 5 updates \\
        \midrule
        Directional-progress clip & 1.0 & $\alpha_{\mathrm{low}}$ / $\alpha_{\mathrm{copy}}$ & 1.0 / 2.0 \\
        Similarity interval & $[\delta_{\mathrm{low}},\delta_{\mathrm{high}})$ & $w_{\mathrm{prop}}$ / $w_{\mathrm{sim}}$ & 1.0 / 1.5 \\
        Early stopping & Joint property/similarity success & $b_{\mathrm{succ}}$ & 0.5 \\
        $r_{\mathrm{inv}}$ / $\lambda_{\mathrm{len}}$ & $-1.0$ / 0.2 & $\ell_{\max}$ & 256 \\
        $\lambda_{\mathrm{imp}}$ / $\lambda_{\mathrm{reg}}$ & 0.5 / 0.75 & $\eta_{\mathrm{imp}}$ / $\eta_{\mathrm{reg}}$ & 0.02 / 0.02 \\
        \bottomrule
    \end{tabular*}
    \caption{Canonical MARCO training, evaluation, and reward settings. Reward coefficients correspond to the turn-reward equations in the main paper; the similarity bounds remain $\delta_{\mathrm{low}}$ and $\delta_{\mathrm{high}}$.}
    \label{tab:training_evaluation_settings}
\end{table}

\section{Feedback-RL Procedure}

Algorithm~\ref{alg:feedback_rl_procedure} summarizes the SFT-initialized multi-turn rollout and trajectory-level update used by \marco{}. Each instruction produces a group of trajectories from the same source molecule. At every turn, the verifier evaluates validity, requested property directions, and source similarity, then appends structured feedback to the history. A trajectory stops after joint property and accepted fingerprint-similarity success or continues to the horizon.

\begin{algorithm}[tb]
\caption{SFT-initialized multi-turn feedback RL}
\label{alg:feedback_rl_procedure}
\small
\textbf{Input}: Instruction $I$, SFT policy $\pi_\theta$, reference $\pi_{\mathrm{ref}}$, group size $K$, horizon $T$\\
\begin{algorithmic}[1]
\STATE Sample $K$ trajectories for instruction $I$
\FOR{each trajectory $\tau_g$}
    \STATE Initialize source $x_0$ and feedback history $h_0 \leftarrow \varnothing$
    \FOR{$t=1$ to $T$}
        \STATE Generate an action from $\pi_\theta(\cdot\mid I,x_0,h_{t-1})$ and parse its candidate
        \IF{the response does not yield an evaluable candidate}
            \STATE Set the candidate to the invalid sentinel; set $r_t$ to the fixed invalid-action reward and append invalid feedback
        \ELSE
            \STATE Evaluate properties and source similarity; construct $r_t$ and append verifier feedback
            \STATE Stop early if the candidate satisfies the joint property and similarity conditions
        \ENDIF
    \ENDFOR
    \STATE Construct $R(\tau_g)=\sum_{t=1}^{T_{\tau_g}}r_t$ from the realized turns
\ENDFOR
\STATE Compute group-relative advantages and update $\theta$, optionally KL-regularized to $\pi_{\mathrm{ref}}$
\end{algorithmic}
\end{algorithm}

Let $r_t$ denote the shaped reward at turn $t$. The trend contribution is zero at $t=1$; the existing improvement and regression terms apply from the second turn onward. For each realized trajectory $\tau$, we construct the undiscounted trajectory return
\begin{equation}
R(\tau)=\sum_{t=1}^{T_\tau} r_t,
\end{equation}
where $T_\tau$ is the number of realized turns. Unexecuted turns are outside the realized trajectory and do not appear in the sum. A response that does not yield an evaluable molecular candidate is represented by an invalid sentinel. This action receives the fixed invalid-action reward, appends invalid-candidate feedback, and continues the feedback trajectory; property deltas, similarity, and valid-candidate quality are not evaluated for the sentinel. For an invalid model action, \marco{} assigns the fixed invalid-action penalty and carries forward the previous state-quality value for trajectory bookkeeping; the invalid turn contributes no improvement or regression term. Subsequent valid turns compare their quality against this carried-forward history.

\section{Complete Ablation Results}

The complete three-backbone Same-5 component table is provided below. We additionally expand the initialization study, four-objective results, and the \gellmoPSixMistral{} public-checkpoint adaptation experiment, reporting SR and Sim alongside SR$\times$Sim.

\subsection{Same-5 Component Results}

The main paper reports the compact Same-5 SR$\times$Sim comparison. The table below provides the corresponding SR and Sim components for all three backbones under the same matched five-response protocol.

\begin{table}[!htbp]
    \centering
    \small
    \setlength{\tabcolsep}{0.75pt}
    \renewcommand{\arraystretch}{1.02}
    \begin{threeparttable}
    \begin{adjustbox}{max width=\linewidth}
    \begin{tabular}{@{}llcccccccccccccccccc@{}}
        \toprule
        \textbf{Metric} & \textbf{Method} & \multicolumn{6}{c}{\qwenThreeBHead} & \multicolumn{6}{c}{\qwenSevenBHead} & \multicolumn{6}{c}{\qwenFourBHead} \\
        \cmidrule(lr){3-8} \cmidrule(lr){9-14} \cmidrule(lr){15-20}
        & & \multicolumn{3}{c}{\textbf{Seen}} & \multicolumn{3}{c}{\textbf{Unseen}} & \multicolumn{3}{c}{\textbf{Seen}} & \multicolumn{3}{c}{\textbf{Unseen}} & \multicolumn{3}{c}{\textbf{Seen}} & \multicolumn{3}{c}{\textbf{Unseen}} \\
        \cmidrule(lr){3-5} \cmidrule(lr){6-8} \cmidrule(lr){9-11} \cmidrule(lr){12-14} \cmidrule(lr){15-17} \cmidrule(lr){18-20}
        & & BDP & BDQ & BPQ & BDP & BDQ & BPQ & BDP & BDQ & BPQ & BDP & BDQ & BPQ & BDP & BDQ & BPQ & BDP & BDQ & BPQ \\
        \midrule
        \multirow{5}{*}{SR} & Base & 0.152 & 0.108 & 0.250 & 0.092 & 0.140 & 0.204 & 0.192 & 0.202 & 0.280 & 0.166 & 0.190 & 0.348 & 0.160 & 0.150 & 0.212 & 0.172 & 0.206 & 0.212 \\
        & SFT & \underline{0.504} & \underline{0.334} & \bnum{0.544} & \underline{0.442} & \underline{0.332} & \bnum{0.578} & \underline{0.408} & 0.344 & \underline{0.616} & \underline{0.446} & 0.376 & \underline{0.618} & \underline{0.480} & \underline{0.300} & \underline{0.534} & \underline{0.468} & \underline{0.282} & \underline{0.558} \\
        & \tabgrpo & 0.114 & 0.072 & 0.186 & 0.118 & 0.078 & 0.184 & 0.154 & 0.098 & 0.294 & 0.168 & 0.120 & 0.294 & 0.002 & 0.000 & 0.344 & 0.000 & 0.000 & 0.328 \\
        & \tabrepo & 0.176 & 0.228 & 0.252 & 0.106 & 0.190 & 0.166 & 0.374 & \underline{0.416} & 0.430 & 0.404 & \underline{0.416} & 0.368 & 0.424 & 0.282 & 0.356 & 0.380 & 0.280 & 0.402 \\
        & \tabmarcostar & \bnum{0.508} & \bnum{0.362} & \underline{0.524} & \bnum{0.448} & \bnum{0.352} & \underline{0.550} & \bnum{0.496} & \bnum{0.610} & \bnum{0.752} & \bnum{0.506} & \bnum{0.572} & \bnum{0.736} & \bnum{0.558} & \bnum{0.306} & \bnum{0.582} & \bnum{0.512} & \bnum{0.316} & \bnum{0.602} \\
        \cmidrule(lr){1-20}
        \multirow{5}{*}{Sim} & Base & 0.629 & 0.700 & 0.564 & 0.668 & 0.683 & 0.647 & \underline{0.666} & \underline{0.672} & \underline{0.681} & \underline{0.683} & \underline{0.685} & 0.619 & \underline{0.677} & 0.626 & \underline{0.654} & \underline{0.665} & 0.691 & \underline{0.703} \\
        & SFT & 0.448 & 0.572 & 0.531 & 0.447 & 0.550 & 0.531 & 0.455 & 0.653 & 0.458 & 0.444 & 0.633 & 0.425 & 0.539 & 0.643 & 0.525 & 0.548 & 0.654 & 0.503 \\
        & \tabgrpo & \underline{0.798} & \bnum{0.920} & \underline{0.857} & \underline{0.809} & \bnum{0.906} & \underline{0.843} & \bnum{0.682} & \bnum{0.811} & \bnum{0.749} & \bnum{0.722} & \bnum{0.825} & \bnum{0.724} & \bnum{0.993} & \bnum{1.000} & \bnum{0.860} & \bnum{0.994} & \bnum{1.000} & \bnum{0.863} \\
        & \tabrepo & \bnum{0.888} & \underline{0.868} & \bnum{0.928} & \bnum{0.938} & \underline{0.892} & \bnum{0.960} & 0.615 & 0.613 & 0.675 & 0.566 & 0.579 & \underline{0.676} & 0.648 & \underline{0.702} & 0.635 & 0.640 & \underline{0.698} & 0.600 \\
        & \tabmarcostar & 0.457 & 0.713 & 0.683 & 0.471 & 0.694 & 0.671 & 0.486 & 0.510 & 0.499 & 0.466 & 0.519 & 0.499 & 0.518 & 0.654 & 0.577 & 0.518 & 0.655 & 0.529 \\
        \cmidrule(lr){1-20}
        \multirow{5}{*}{SR$\times$Sim} & Base & 0.096 & 0.076 & 0.141 & 0.061 & 0.096 & 0.132 & 0.128 & 0.136 & 0.191 & 0.113 & 0.130 & 0.215 & 0.108 & 0.094 & 0.139 & 0.114 & 0.142 & 0.149 \\
        & SFT & \underline{0.226} & 0.191 & \underline{0.289} & \underline{0.198} & \underline{0.183} & \underline{0.307} & 0.185 & 0.225 & 0.282 & 0.198 & 0.238 & \underline{0.263} & 0.259 & 0.193 & 0.280 & \underline{0.256} & 0.184 & 0.281 \\
        & \tabgrpo & 0.091 & 0.066 & 0.159 & 0.095 & 0.071 & 0.155 & 0.105 & 0.079 & 0.220 & 0.121 & 0.099 & 0.213 & 0.002 & 0.000 & \underline{0.296} & 0.000 & 0.000 & \underline{0.283} \\
        & \tabrepo & 0.156 & \underline{0.198} & 0.234 & 0.099 & 0.169 & 0.159 & \underline{0.230} & \underline{0.255} & \underline{0.290} & \underline{0.229} & \underline{0.241} & 0.249 & \underline{0.275} & \underline{0.198} & 0.226 & 0.243 & \underline{0.195} & 0.241 \\
        & \tabmarcostar & \bnum{0.232} & \bnum{0.258} & \bnum{0.358} & \bnum{0.211} & \bnum{0.244} & \bnum{0.369} & \bnum{0.241} & \bnum{0.311} & \bnum{0.375} & \bnum{0.236} & \bnum{0.297} & \bnum{0.367} & \bnum{0.289} & \bnum{0.200} & \bnum{0.336} & \bnum{0.265} & \bnum{0.207} & \bnum{0.318} \\
        \bottomrule
    \end{tabular}\end{adjustbox}
    \begin{tablenotes}[flushleft]
        \small
        \item[$*$] Initialization from the corresponding per-subtask SFT checkpoint. Best and second-best values within each metric/backbone/split/objective column are bolded and underlined, respectively.
    \end{tablenotes}
    \caption{Complete Same-5 component results. Backbones are grouped horizontally; rows report SR, Sim, and SR$\times$Sim for seen/unseen instruction splits and the three primary objectives.}
    \label{tab:same5_full_supp}
    \end{threeparttable}
\end{table}

\subsection{Initialization Ablation}

Table~\ref{tab:init_ablation_full} gives all \qwenThreeB{} Same-1 metrics. Starred methods use the corresponding per-subtask SFT initialization. \marco{}\sftinit{} obtains the best SR$\times$Sim in all six settings; GRPO\sftinit{} retains high similarity but lower success, and unstarred \marco{} improves over the base model without matching the combined recipe.

\begin{table}[!htbp]
    \centering
    \small
    \setlength{\tabcolsep}{3.4pt}
    \renewcommand{\arraystretch}{1.02}
    \begin{threeparttable}
    \begin{tabular*}{\linewidth}{@{\extracolsep{\fill}}llcccccc@{}}
        \toprule
        \textbf{Metric} & \textbf{Method} & \multicolumn{3}{c}{\textbf{Seen instruction}} & \multicolumn{3}{c}{\textbf{Unseen instruction}} \\
        \cmidrule(lr){3-5} \cmidrule(lr){6-8}
        & & \textbf{BDP} & \textbf{BDQ} & \textbf{BPQ} & \textbf{BDP} & \textbf{BDQ} & \textbf{BPQ} \\
        \midrule
        \multirow{7}{*}{SR}
        & Base & 0.052 & 0.034 & 0.052 & 0.052 & 0.042 & 0.050 \\
        & SFT & \underline{0.398} & \bnum{0.319} & \bnum{0.471} & \underline{0.310} & \bnum{0.342} & \underline{0.419} \\
        & \tabgrpo & 0.156 & 0.082 & 0.212 & 0.148 & 0.078 & 0.186 \\
        & \tabgrpostar & 0.090 & 0.128 & 0.160 & 0.072 & 0.138 & 0.186 \\
        & \tabrepo & 0.206 & 0.160 & 0.274 & 0.198 & 0.170 & 0.242 \\
        & \tabmarco & 0.184 & 0.086 & 0.124 & 0.130 & 0.086 & 0.178 \\
        & \tabmarcostar & \bnum{0.418} & \underline{0.288} & \underline{0.440} & \bnum{0.378} & \underline{0.274} & \bnum{0.512} \\
        \cmidrule(lr){1-8}
        \multirow{7}{*}{Sim}
        & Base & 0.149 & 0.117 & 0.194 & 0.143 & 0.104 & 0.130 \\
        & SFT & 0.254 & 0.279 & 0.244 & 0.261 & 0.257 & 0.248 \\
        & \tabgrpo & 0.759 & 0.479 & 0.567 & 0.727 & \underline{0.457} & 0.573 \\
        & \tabgrpostar & \bnum{0.911} & \bnum{0.782} & \bnum{0.961} & \bnum{0.922} & \bnum{0.781} & \bnum{0.944} \\
        & \tabrepo & 0.569 & 0.365 & 0.509 & 0.572 & 0.322 & 0.596 \\
        & \tabmarco & \underline{0.775} & \underline{0.761} & \underline{0.737} & \underline{0.812} & \bnum{0.781} & \underline{0.659} \\
        & \tabmarcostar & 0.485 & 0.576 & 0.527 & 0.464 & 0.570 & 0.523 \\
        \cmidrule(lr){1-8}
        \multirow{7}{*}{SR$\times$Sim}
        & Base & 0.008 & 0.004 & 0.010 & 0.007 & 0.004 & 0.007 \\
        & SFT & 0.101 & 0.089 & 0.115 & 0.081 & 0.088 & 0.104 \\
        & \tabgrpo & 0.118 & 0.039 & 0.120 & 0.108 & 0.036 & 0.107 \\
        & \tabgrpostar & 0.082 & \underline{0.100} & \underline{0.154} & 0.066 & \underline{0.108} & \underline{0.176} \\
        & \tabrepo & 0.117 & 0.058 & 0.139 & \underline{0.113} & 0.055 & 0.144 \\
        & \tabmarco & \underline{0.143} & 0.065 & 0.091 & 0.106 & 0.067 & 0.117 \\
        & \tabmarcostar & \bnum{0.203} & \bnum{0.166} & \bnum{0.232} & \bnum{0.175} & \bnum{0.156} & \bnum{0.268} \\
        \bottomrule
    \end{tabular*}
    \begin{tablenotes}[flushleft]
        \small
        \item[$*$] Initialization from the corresponding per-subtask SFT checkpoint.
        \item[] Best and second-best values in each metric/objective/split column are bolded and underlined. Tied best values are both bolded.
    \end{tablenotes}
    \caption{Complete SFT-initialization ablation on \qwenThreeB{} under Same-1.}
    \label{tab:init_ablation_full}
    \end{threeparttable}
\end{table}

\subsection{Candidate-Level SWS Details}
\label{sec:candidate_level_similarity}

The main paper reports SWS for all three backbones. SWS uses the complete evaluation split as a fixed denominator: invalid candidates, candidates that fail any requested property direction, and exact source copies contribute zero. We impose no additional similarity threshold. Exact molecular identity is determined using canonical isomeric SMILES with stereochemistry retained; it is separate from the fingerprint-similarity interval $[\delta_{\mathrm{low}},\delta_{\mathrm{high}})$.

For the \qwenSevenB{} audit, GRPO produces 237, 204, 112, 50, 141, and 49 exact source copies for BDP seen/unseen, BDQ seen/unseen, and BPQ seen/unseen, respectively. Among these, 39, 37, 13, 10, 18, and 8 satisfy the uncorrected strict-positive property-success rule through numerical predictor variation. RePO has 12, 0, 6, 0, 5, and 0 property-success exact copies, whereas Base, SFT, and MARCO have none. These copies receive zero contribution in SWS.

\subsection{Four-Objective Extension}

The four-objective extension increases the number of simultaneously requested directions from three to four and is evaluated on \qwenSevenB{}. Table~\ref{tab:four_obj_full} reports SR, Sim, and SR$\times$Sim for both evaluation budgets. SFT-initialized MARCO has the highest SR and SR$\times$Sim for HMPQ and BDPQ on both instruction splits.

\subsection{Public-Checkpoint Adaptation}

Table~\ref{tab:p6_full} in the main text reports public-checkpoint adaptation with its component metrics. The starting point is the released \gellmoPSixMistral{} checkpoint built on \mistralSevenB{} \cite{dey-etal-2025-gellm3o,jiang2023mistral}. Feedback RL improves property success and SR$\times$Sim in all six rows. Similarity increases in four rows, while the released checkpoint has higher similarity on the two BDQ rows.

\begin{table}[!htbp]
    \centering
    \small
    \setlength{\tabcolsep}{2.4pt}
    \renewcommand{\arraystretch}{0.98}
    \begin{threeparttable}
    \begin{tabular*}{\linewidth}{@{\extracolsep{\fill}}llcccccccc@{}}
        \toprule
        \textbf{Metric} & \textbf{Method} & \multicolumn{4}{c}{\textbf{Same-1}} & \multicolumn{4}{c}{\textbf{Same-5}} \\
        \cmidrule(lr){3-6} \cmidrule(lr){7-10}
        & & \multicolumn{2}{c}{\textbf{HMPQ}} & \multicolumn{2}{c}{\textbf{BDPQ}} & \multicolumn{2}{c}{\textbf{HMPQ}} & \multicolumn{2}{c}{\textbf{BDPQ}} \\
        \cmidrule(lr){3-4} \cmidrule(lr){5-6} \cmidrule(lr){7-8} \cmidrule(lr){9-10}
        & & \textbf{Seen} & \textbf{Unseen} & \textbf{Seen} & \textbf{Unseen} & \textbf{Seen} & \textbf{Unseen} & \textbf{Seen} & \textbf{Unseen} \\
        \midrule
        \multirow{5}{*}{SR}
        & Base & 0.042 & 0.062 & 0.024 & 0.040 & 0.146 & 0.146 & 0.054 & 0.082 \\
        & SFT & \underline{0.385} & \underline{0.469} & \underline{0.202} & \underline{0.186} & \underline{0.469} & \underline{0.510} & \underline{0.252} & \underline{0.232} \\
        & \tabgrpo & 0.042 & 0.021 & 0.068 & 0.040 & 0.167 & 0.104 & 0.096 & 0.058 \\
        & \tabrepo & 0.135 & 0.156 & 0.090 & 0.068 & 0.156 & 0.198 & 0.138 & 0.084 \\
        & \tabmarcostar & \bnum{0.458} & \bnum{0.510} & \bnum{0.222} & \bnum{0.210} & \bnum{0.562} & \bnum{0.552} & \bnum{0.336} & \bnum{0.324} \\
        \cmidrule(lr){1-10}
        \multirow{5}{*}{Sim}
        & Base & 0.634 & 0.597 & \underline{0.683} & \underline{0.672} & \underline{0.682} & 0.605 & \underline{0.698} & \underline{0.700} \\
        & SFT & 0.577 & 0.534 & 0.631 & 0.626 & 0.547 & 0.528 & 0.605 & 0.608 \\
        & \tabgrpo & \underline{0.682} & \underline{0.654} & \bnum{0.788} & \bnum{0.780} & 0.669 & \underline{0.714} & \bnum{0.827} & \bnum{0.839} \\
        & \tabrepo & \bnum{0.715} & \bnum{0.730} & 0.641 & 0.630 & \bnum{0.744} & \bnum{0.747} & 0.592 & 0.577 \\
        & \tabmarcostar & 0.562 & 0.557 & 0.605 & 0.617 & 0.525 & 0.539 & 0.493 & 0.490 \\
        \cmidrule(lr){1-10}
        \multirow{5}{*}{SR$\times$Sim}
        & Base & 0.026 & 0.037 & 0.016 & 0.027 & 0.099 & 0.088 & 0.038 & 0.057 \\
        & SFT & \underline{0.222} & \underline{0.250} & \underline{0.127} & \underline{0.117} & \underline{0.256} & \underline{0.269} & \underline{0.152} & \underline{0.141} \\
        & \tabgrpo & 0.028 & 0.014 & 0.054 & 0.031 & 0.111 & 0.074 & 0.079 & 0.049 \\
        & \tabrepo & 0.097 & 0.114 & 0.058 & 0.043 & 0.116 & 0.148 & 0.082 & 0.048 \\
        & \tabmarcostar & \bnum{0.257} & \bnum{0.285} & \bnum{0.134} & \bnum{0.130} & \bnum{0.296} & \bnum{0.298} & \bnum{0.166} & \bnum{0.159} \\
        \bottomrule
    \end{tabular*}
    \vspace{4pt}

    \begin{tablenotes}[flushleft]
        \small
        \item[$*$] Initialization from the corresponding per-subtask SFT checkpoint.
    \end{tablenotes}
    \caption{Complete four-objective Same-1 and Same-5 results on \qwenSevenB{}.}
    \label{tab:four_obj_full}
    \end{threeparttable}
\end{table}

\section{Prompt and Feedback Templates}

The templates use semantic placeholders for the policy--verifier interaction. Candidate molecules are parsed from a single \texttt{<SMILES>...\allowbreak</SMILES>} block.

\vspace{4pt}

\begin{supplementtext}{Policy and instruction templates}
\textbf{System.} You are a molecular optimization assistant. Reason concisely about the requested edits, then return exactly one valid, connected molecule enclosed in \texttt{<SMILES>} and \texttt{</SMILES>}. Do not return multiple candidates or text after the closing tag.

\textbf{User.} Source molecule: \texttt{\{SOURCE\}}. Modify it to \texttt{\{DIRECTION\}} each property in \texttt{\{PROPERTY SET\}} while retaining the core structure as much as possible.

\textbf{Output.} \texttt{<SMILES>\{CANDIDATE\}</SMILES>}
\end{supplementtext}

\begin{supplementtext}{Revision-context template}
Previous candidate: \texttt{\{CANDIDATE\}}. Predictor and similarity feedback: \texttt{\{FEEDBACK\}}. Propose one revised molecule that addresses the remaining misses without unnecessary scaffold change.
\end{supplementtext}

\begin{supplementtext}{Valid-candidate feedback}
Candidate is valid. For each active property: predicted value \texttt{\{VALUE\}}, requested direction \texttt{\{DIRECTION\}}, directional change \texttt{\{DELTA\}}, and progress status \texttt{\{STATUS\}}. Similarity to the source is \texttt{\{SIMILARITY\}} with respect to the target scaffold-preservation range. All requested property directions satisfied: \texttt{\{YES/NO\}}. Revise only the remaining misses.
\end{supplementtext}

\begin{supplementtext}{Invalid-candidate feedback}
Candidate is invalid: \texttt{\{ERROR TYPE\}}. Reason: \texttt{\{ERROR MESSAGE\}}. Return exactly one connected, parseable molecule. Preserve the source scaffold where possible and avoid disconnected fragments or multiple alternatives.
\end{supplementtext}

\begin{supplementtext}{Similarity-range feedback}
\textbf{Near copy.} Candidate is too close to the source to receive full editing credit. Make a targeted structural change that advances the requested properties while remaining within the accepted similarity region.

\textbf{Scaffold drift.} Candidate has moved outside the accepted scaffold-preservation region. Restore more of the source scaffold while retaining property-improving edits.
\end{supplementtext}

\section{Qualitative Trajectory Details}

Figure~2 in the main paper summarizes an illustrative \qwenThreeB{} SFT-initialized MARCO BPQ trajectory using edited local fragments, with $R$ denoting omitted structure. It reports signed property changes---green when the requested direction is met and vermillion otherwise---together with source similarity and the number of goals met at each turn. We select this case because it spans all five turns, contains intermediate candidates that miss different requested directions, and ends with a property-success candidate. The first four candidates miss at least one requested direction; Turn~5 satisfies all three at similarity 0.627.

Figure~\ref{fig:case_study_details} provides the complementary full-trajectory view: it separates the three property scales, marks the directional-success boundary at zero, reports source similarity on the shared secondary scale, and shows the complete candidate molecules rather than only the edited local fragments.

% Keep the final case-study paragraph together above the figure page.
\enlargethispage{\baselineskip}
\subsection{Qualitative Case Study}

\begin{figure}[!t]
    \centering
    \includegraphics[width=0.50\textwidth,height=0.54\textheight,keepaspectratio]{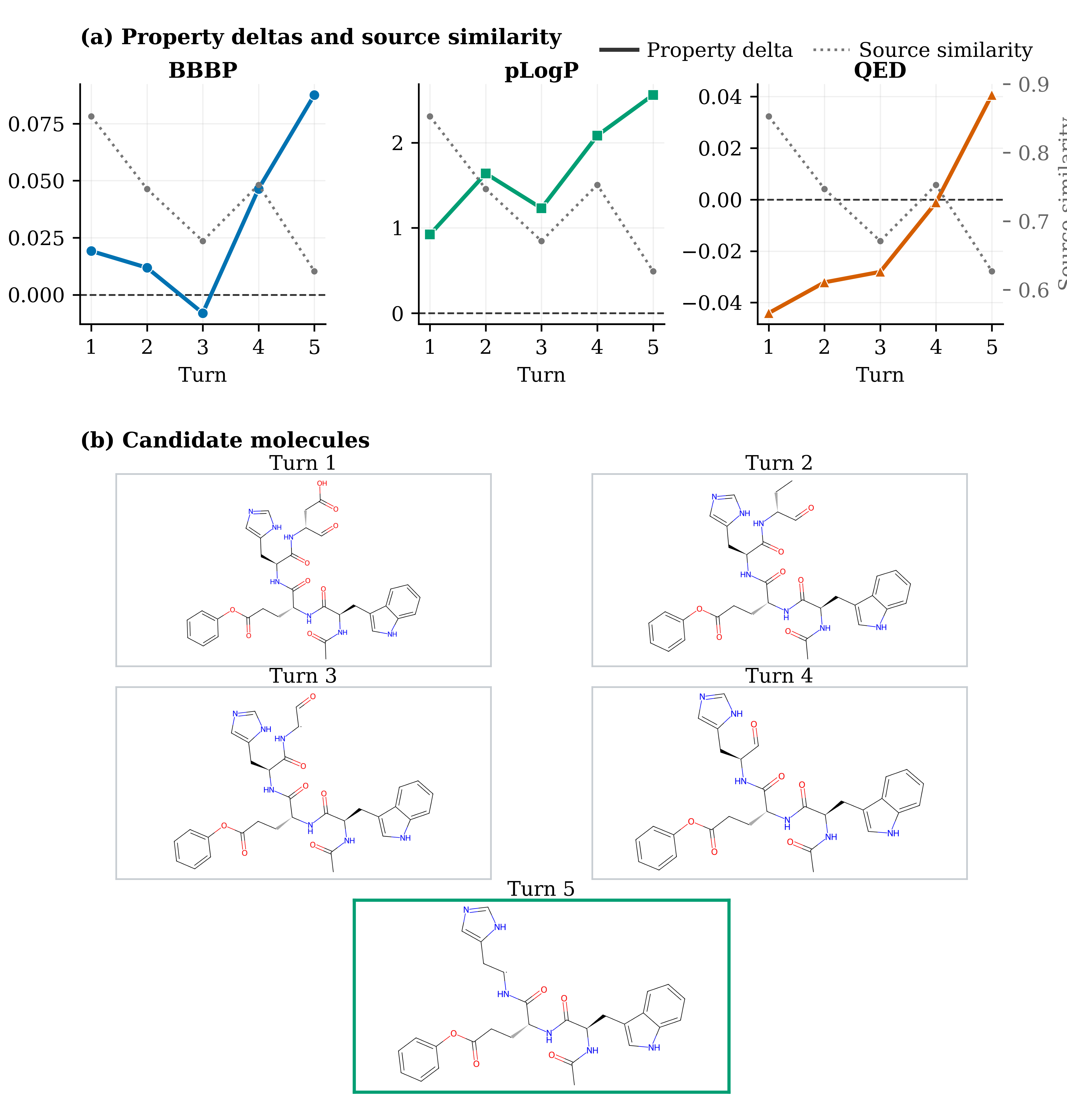}
    \caption{Complete view of the illustrative \qwenThreeB{} SFT-initialized MARCO BPQ trajectory summarized by Figure~2 in the main paper. Panel (a) shows directional BBBP, pLogP, and QED changes with source similarity; panel (b) shows the five complete RDKit-rendered candidates. The green frame marks the first property-success candidate at Turn~5.}
    \label{fig:case_study_details}
    \vspace{4pt}
\end{figure}

The five-turn trajectory in Figure~\ref{fig:case_study_details} illustrates successive proposals that miss different requested property directions before the final candidate achieves joint improvement in BBBP, penalized logP, and QED with a source similarity of 0.627. This example serves only as a trajectory-level illustration of feedback use rather than a substitute for the aggregate evaluation; the source molecule, candidate SMILES, RDKit depictions, and raw predictor deltas are provided below.

\begin{supplementtext}{Source instruction and molecule}
\textbf{Source instruction.} Adjust the structure of the given molecule to target the specified adjustments in molecular properties. Retain the core structure as much as possible. Respond with only the SMILES of the modified molecule enclosed in \texttt{<SMILES>} \texttt{</SMILES>} tags.

\textbf{Source SMILES (Input).}
\texttt{<SMILES>}\url{CC(=O)N[C@H](Cc1c[nH]c2ccccc12)C(=O)N[C@H](CCC(=O)O)C(=O)N[C@@H](Cc1cnc[nH]1)C(=O)N[C@@H](C=O)CC(=O)O}\texttt{</SMILES>}

\textbf{Requested adjustment.} increase BBB permeability, increase Penalized octanol-water partition coefficient (penalized logP) and increase QED
\end{supplementtext}

\begin{supplementtext}{Five-turn candidate sequence}
\textbf{Turn 1.}
\texttt{<SMILES>}\url{CC(=O)N[C@H](Cc1c[nH]c2ccccc12)C(=O)N[C@H](CCC(=O)Oc3ccccc3)C(=O)N[C@@H](Cc1cnc[nH]1)C(=O)N[C@@H](C=O)CC(=O)O}\texttt{</SMILES>}

\textbf{Turn 2.}
\texttt{<SMILES>}\url{CC(=O)N[C@H](Cc1c[nH]c2ccccc12)C(=O)N[C@H](CCC(=O)Oc3ccccc3)C(=O)N[C@@H](Cc1cnc[nH]1)C(=O)N[C@@H](C=O)CC}\texttt{</SMILES>}

\textbf{Turn 3.}
\texttt{<SMILES>}\url{CC(=O)N[C@H](Cc1c[nH]c2ccccc12)C(=O)N[C@H](CCC(=O)Oc3ccccc3)C(=O)N[C@@H](Cc1cnc[nH]1)C(=O)N[C@@H](C=O)}\texttt{</SMILES>}

\textbf{Turn 4.}
\texttt{<SMILES>}\url{CC(=O)N[C@H](Cc1c[nH]c2ccccc12)C(=O)N[C@H](CCC(=O)Oc3ccccc3)C(=O)N[C@@H](Cc1cnc[nH]1)C(=O)}\texttt{</SMILES>}

\textbf{Turn 5.}
\texttt{<SMILES>}\url{CC(=O)N[C@H](Cc1c[nH]c2ccccc12)C(=O)N[C@H](CCC(=O)Oc3ccccc3)C(=O)N[C@@H](Cc1cnc[nH]1)}\texttt{</SMILES>}
\end{supplementtext}

\end{document}